\documentclass[12pt]{article}

\usepackage[margin=1in]{geometry}
\usepackage[utf8]{inputenc}
\usepackage[T1]{fontenc}
\usepackage{hyperref}
\hypersetup{colorlinks=true, linkcolor=blue, citecolor=blue, urlcolor=blue}
\usepackage{url}
\usepackage{booktabs}
\usepackage{tabularx}
\usepackage{amsfonts}
\usepackage{amsmath}
\usepackage{amssymb}
\usepackage{nicefrac}
\usepackage{microtype}
\usepackage{xcolor}
\usepackage{graphicx}
\usepackage{subcaption}
\usepackage{multirow}
\usepackage{algorithm}
\usepackage{algpseudocode}
\usepackage{natbib}
\usepackage{xspace}

\newcommand{\TPW}{\textit{Territory Paint Wars}}

\newcommand{\realfig}[3][6cm]{%
  \includegraphics[width=\linewidth]{#2}%
}
\graphicspath{{figures/}}

\title{%
  Territory Paint Wars: Diagnosing and Mitigating\\
  Failure Modes in Competitive Multi-Agent PPO
}

\author{%
  Diyansha Singh\\
  Independent Researcher\\
  \texttt{diyansha.singh@gmail.com}
}

\date{}

\begin{document}
\pdfpageattr{/Group << /S /Transparency /I true /CS /DeviceRGB>>}
\maketitle

\begin{abstract}
We present \TPW{}, a minimal competitive multi-agent reinforcement learning
environment implemented in Unity, and use it to systematically investigate
failure modes of Proximal Policy Optimisation (PPO) under self-play.
A first agent trained for 84{,}000 episodes achieves only 26.8\% win rate
against a uniformly-random opponent in a symmetric zero-sum game.
Through controlled ablations we identify five implementation-level failure
modes---reward-scale imbalance, missing terminal signal, ineffective
long-horizon credit assignment, unnormalised observations, and incorrect win
detection---each of which contributes critically to this failure in this setting.

After correcting these issues, we uncover a distinct emergent pathology:
\textbf{competitive overfitting}, where co-adapting agents maintain stable
self-play performance while generalisation win rate collapses from 73.5\% to
21.6\%.
Critically, this failure is \emph{undetectable} via standard self-play
metrics: both agents co-adapt equally, so the self-play win rate remains near
50\% throughout the collapse.

We propose a minimal intervention---\emph{opponent mixing}, where 20\% of
training episodes substitute a fixed uniformly-random policy for the
co-adaptive opponent---which mitigates competitive overfitting and restores
generalisation to $77.1\%$ ($\pm 12.6\%$, 10 seeds) without population-based
training or additional infrastructure.
Our results demonstrate that in this competitive setting,
self-play alone is insufficient for robust generalisation, and that
maintaining opponent diversity is essential.
We open-source \TPW{} to provide a reproducible benchmark for studying
competitive MARL failure modes.
\end{abstract}

\section{Introduction}
\label{sec:intro}

Competitive multi-agent reinforcement learning (MARL) is an active research
area motivated by the observation that self-play can produce agents that
surpass human performance in complex strategy games
\citep{silver2017mastering,vinyals2019grandmaster,berner2019dota}.
Yet practitioners frequently report that standard single-agent RL algorithms
transplanted to multi-agent settings fail to learn even simple competitive
tasks \citep{lowe2017multi}.
The reasons are often diffuse---reward misspecification, credit assignment
difficulties, and training instability each contribute---making it hard to
attribute failure to any single cause.

This paper makes failure causes concrete.
We build \TPW{}, a deterministic, zero-sum two-player grid game implemented in
Unity with a custom Python--Unity TCP bridge, and document precisely
\emph{why} vanilla PPO underperforms in this setting and \emph{which fixes
are individually critical}.

\paragraph{The game.}
\TPW{} is played on a $10{\times}10$ grid.
Two agents (Pink and Green) begin at symmetric starting positions and act
simultaneously each step.
An agent may move in one of four cardinal directions or \emph{lock} the tile
it currently occupies; a locked tile cannot be reclaimed by the opponent.
The agent controlling the most tiles after 250 steps wins.
The irreversible lock mechanic introduces genuine commitment decisions, creating
richer strategic depth than pure territory-painting games.

\begin{figure}[t]
  \centering
  \begin{subfigure}[t]{0.32\textwidth}
    \includegraphics[width=\linewidth]{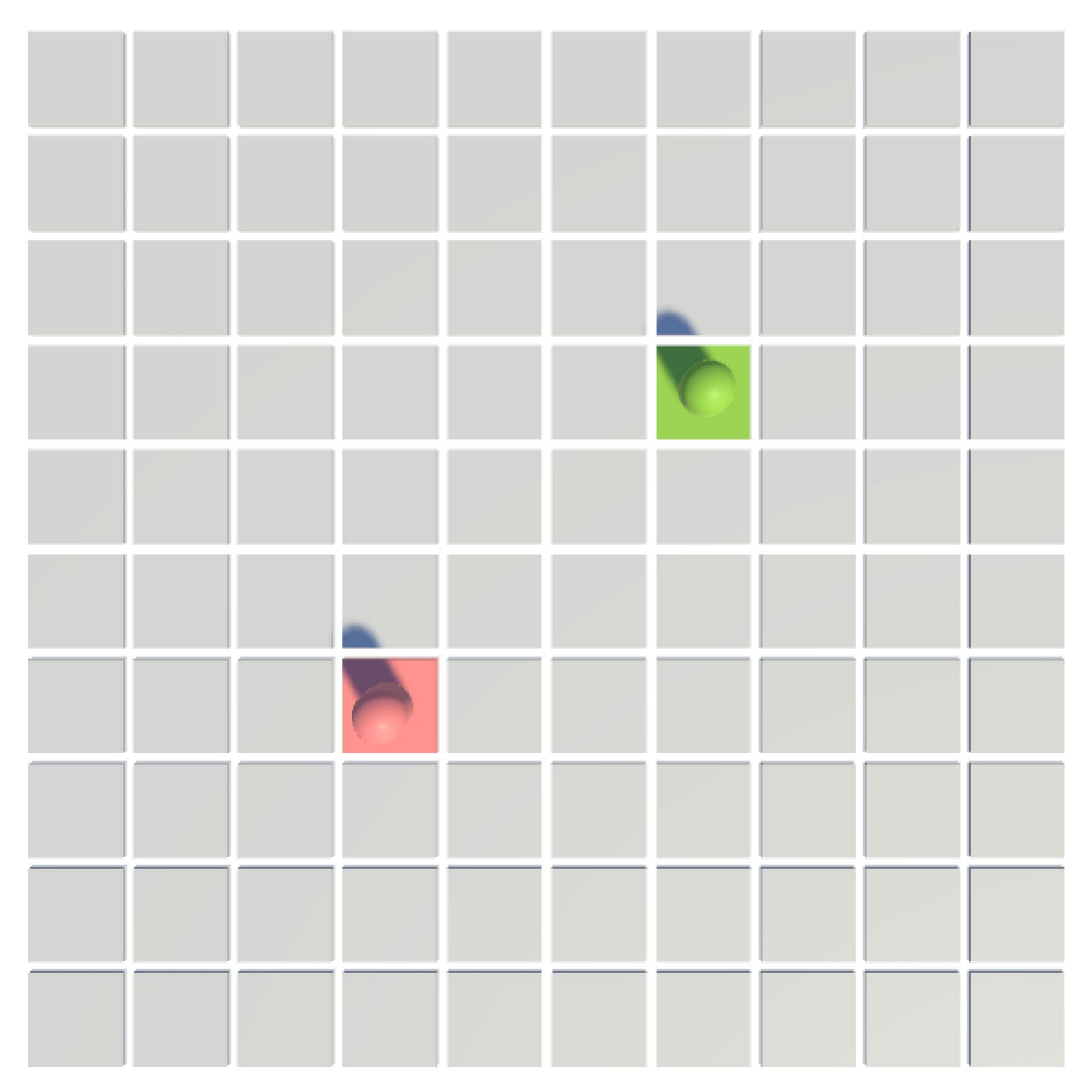}
    \caption*{Start (step 0)}
  \end{subfigure}\hfill
  \begin{subfigure}[t]{0.32\textwidth}
    \includegraphics[width=\linewidth]{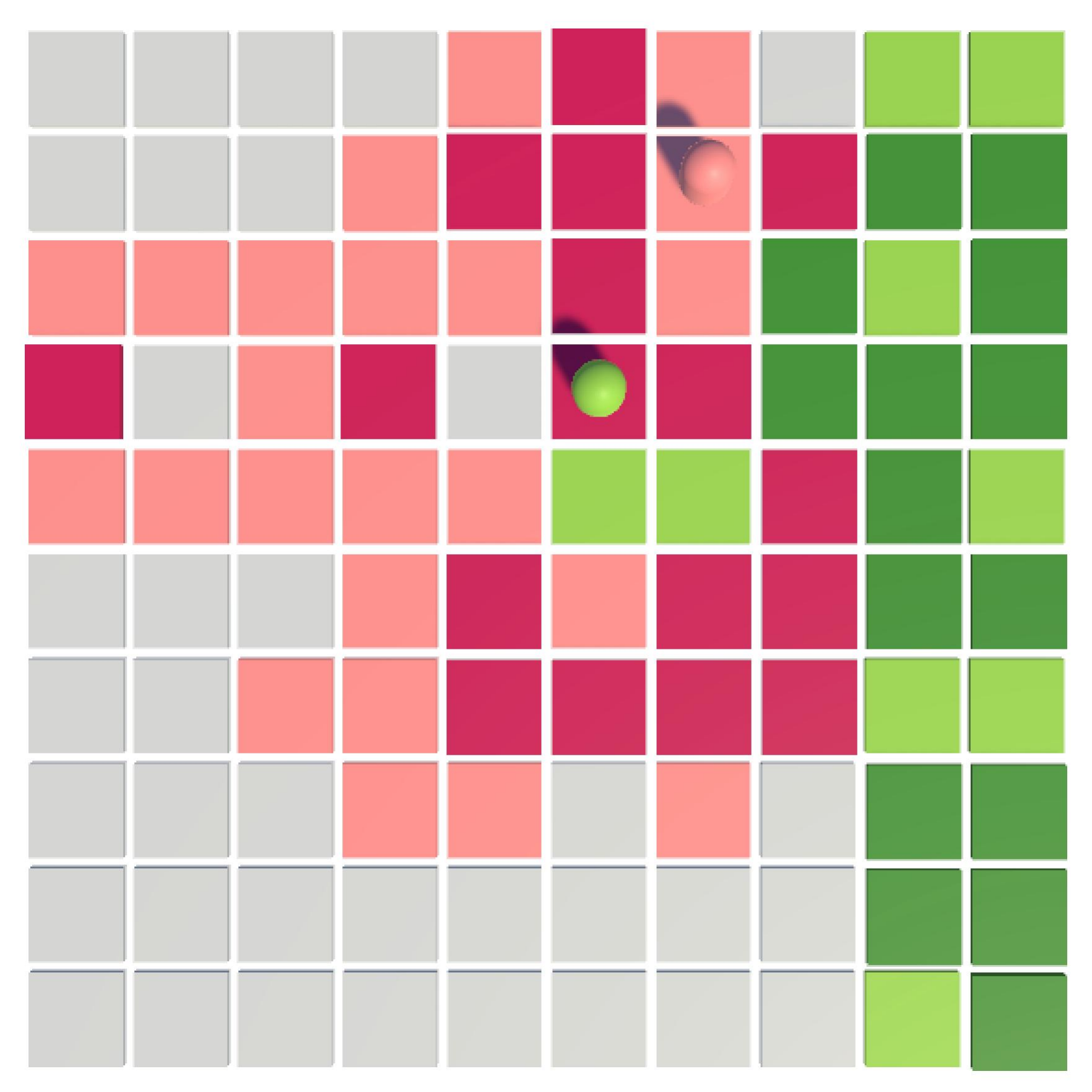}
    \caption*{Mid-game (step 125)}
  \end{subfigure}\hfill
  \begin{subfigure}[t]{0.32\textwidth}
    \includegraphics[width=\linewidth]{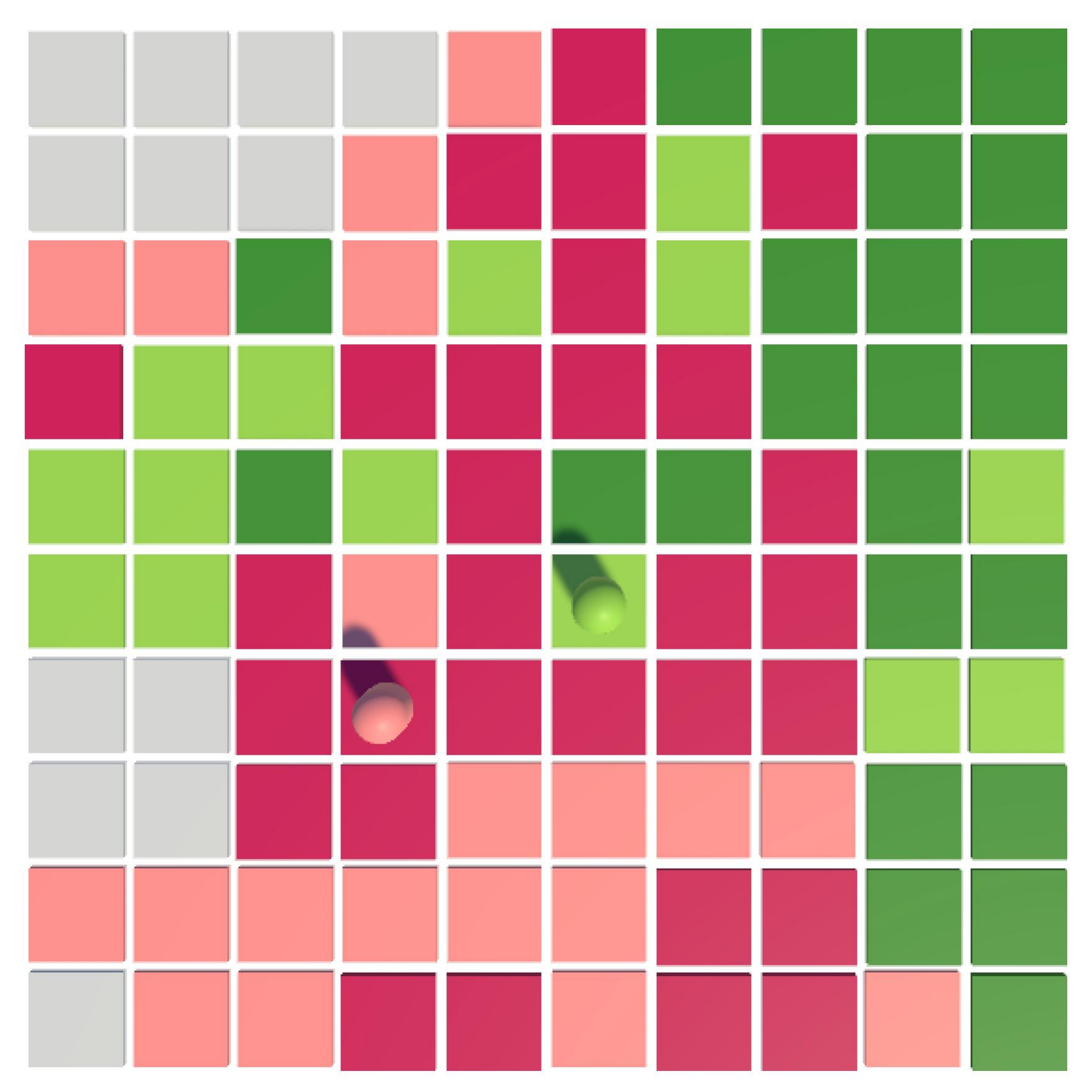}
    \caption*{Final (step 250)}
  \end{subfigure}
  \caption{The \TPW{} environment on a $10{\times}10$ grid.
    Light pink/green tiles are owned by each agent; dark pink/dark green tiles
    are \emph{locked} and cannot be reclaimed; grey tiles are neutral.
    Agents (spheres) paint the tile they occupy each step.}
  \label{fig:env}
\end{figure}

\begin{figure}[t]
  \centering
  \realfig[5cm]{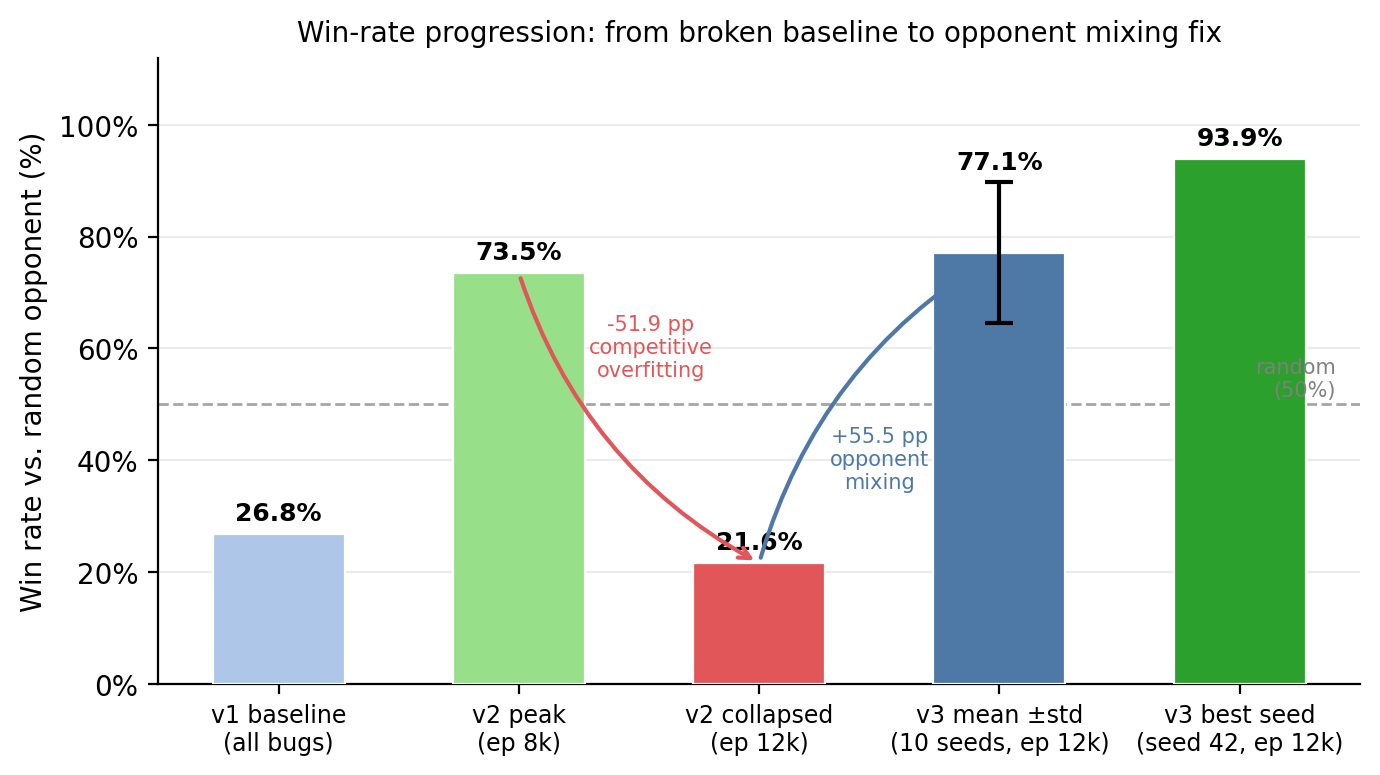}{%
    Bar chart: win rate vs.\ random for v1 (26.8\%), v2 peak ep 8k (73.5\%),
    v2 ep 12k (21.6\%), v3 mean±std (77.1\%±12.6\%, 10 seeds), v3 best (93.9\%).}
  \caption{Win-rate progression across all versions.
    The red bar (v2 at ep 12{,}000) exposes competitive overfitting; the blue bar
    (v3 mean, 10 seeds) shows the recovery via opponent mixing.
    Error bar on v3 mean shows $\pm$1 SD across seeds.}
  \label{fig:progression}
\end{figure}

\paragraph{Contributions.}
\begin{enumerate}
  \item We open-source \TPW{} as a competitive RL benchmark with a
    Unity--Python TCP interface requiring no ML-Agents dependency.
  \item We systematically identify and fix five compounding implementation
    failure modes that together cause a well-tuned PPO agent to perform worse
    than random (26.8\% win rate).
  \item We identify and characterise \textbf{competitive overfitting}: a
    self-play failure mode in which win rate against random opponents collapses
    from 73.5\% to 21.6\% over 4{,}000 episodes of continued training, while
    self-play win rates give no warning signal.
  \item We show that \emph{opponent mixing} ($\varepsilon=0.2$, a single
    one-line code change) mitigates competitive overfitting, recovering mean
    generalisation win rate to $77.1\%$ ($\pm 12.6\%$, 10 seeds).
  \item We isolate each fix's contribution via ablation, yielding a
    discriminative result: GAE, observation normalisation, and opponent
    mixing are each individually critical (removing any one yields
    $\leq 21.6\%$, below the broken v1 baseline), while terminal reward
    is complementary---its removal still yields $87.1\%$, showing that
    dense step rewards alone are sufficient when other fixes are in place.
\end{enumerate}

\section{Related Work}
\label{sec:related}

\paragraph{Competitive multi-agent RL.}
Early work on competitive MARL focused on large-scale games such as
Go \citep{silver2017mastering} and StarCraft II \citep{vinyals2019grandmaster},
where self-play produces increasingly capable agents.
\citet{bansal2018emergent} show that simple competitive environments can produce
complex emergent behaviours with standard PPO, but also observe co-adaptive
instability---consistent with the competitive overfitting we characterise here.
\citet{lowe2017multi} develop MADDPG for cooperative--competitive settings with
centralised training and decentralised execution; unlike their work, we use
fully decentralised training with no shared information between agents.

The closest related observation to competitive overfitting is the
``forgetting'' phenomenon in population-based training
\citep{vinyals2019grandmaster}, where agents trained against a fixed pool can
lose to strategies not in the pool.
Fictitious self-play \citep{heinrich2016deep} and prioritised fictitious
self-play (PFSP, used in AlphaStar) address this by sampling historical
opponent checkpoints; our opponent mixing is a lightweight alternative that
requires no checkpoint storage and no opponent selection policy.
\citet{balduzzi2019open} formalise the cyclic dominance structure underlying
such forgetting using game-theoretic tools (\emph{$\alpha$-rank});
our work provides an empirical, reproducible case study of the same phenomenon
in a minimal two-agent setting where the collapse and recovery are fully
measurable.
We make the \emph{monitoring failure} explicit: the self-play win rate gives
no warning, making periodic external evaluation essential.

\paragraph{Reward shaping and scale.}
\citet{ng1999policy} prove that potential-based reward shaping is
policy-invariant; our tile-gain reward ($+0.1$ per tile claimed) approximates
a potential function over territory control.
\citet{andrychowicz2020learning} conduct a large-scale study of what matters
for on-policy deep actor-critic methods, including reward normalisation and
gradient clipping---both adopted here.
The \emph{reward-scale} failure we document (cumulative lock bonus reaching
$\pm$10{,}000) is distinct from reward misspecification: the sign is correct
but the scale dominates all other signals, a failure mode that is easy to
overlook but catastrophic for learning.

\paragraph{Advantage estimation and credit assignment.}
GAE \citep{schulman2015high} is standard in PPO \citep{schulman2017proximal}
but its \emph{necessity} in long-horizon competitive games has not been
systematically studied.
We show that in 250-step episodes with $\gamma=0.99$, plain Monte Carlo
returns reduce step-1 advantages by a factor of $0.99^{249} \approx 0.08$,
making GAE not merely a convenience in this setting.
\citet{hung2019optimizing} and \citet{raposo2021synthetic} study long-horizon
credit assignment; our work adds a concrete, reproducible competitive case study.

\section{Environment: Territory Paint Wars}
\label{sec:env}

\subsection{Game Mechanics}
\label{sec:mechanics}

\TPW{} is implemented in Unity 2022.3 LTS and communicates with Python via a
custom TCP socket (port 9000) using newline-delimited JSON messages.
This design requires no ML-Agents and works with any Python RL library.

\paragraph{State space.}
Each agent receives a 206-dimensional observation vector:
\begin{itemize}
  \item $[0{:}2]$: Own $(x, z)$ grid position (raw: 0--9).
  \item $[2{:}4]$: Opponent $(x, z)$ grid position (raw: 0--9).
  \item $[4{:}104]$: Board ownership (0=neutral, 1=own, 2=opponent) for all
    100 tiles.
  \item $[104{:}204]$: Lock mask (0 or 1) for all 100 tiles.
  \item $[204]$: Steps remaining (raw: 0--250).
  \item $[205]$: Padding.
\end{itemize}

\paragraph{Action space.}
Five discrete actions: move north, south, east, west, or lock the current tile.

\paragraph{Starting positions and fairness.}
Pink starts at $(3,3)$, Green at $(6,6)$.
Both are equidistant from the centre ($\approx 2.12$ tiles).
We verify fairness empirically: 1{,}000 random-vs-random episodes yield
$50.0 \pm 0.5\%$ win rate for each agent.

\paragraph{Episode termination.}
Episodes end deterministically at step 250.
The agent with the most tiles wins; ties occur in $<2\%$ of episodes.

\subsection{Reward Function}
\label{sec:reward}

\begin{equation}
  r_t^{(i)} = \underbrace{0.1 \cdot \Delta\text{tiles}_t^{(i)}}_{\text{territory gain}}
             + \underbrace{0.5 \cdot \Delta\text{locks}_t^{(i)}}_{\text{lock bonus}}
             + \underbrace{\mathbf{1}[\text{done}] \cdot
               \text{sign}\!\left(\text{tiles}^{(i)} - \text{tiles}^{(-i)}\right)}_{\text{terminal win/loss}}
  \label{eq:reward}
\end{equation}
where $\Delta\text{tiles}_t^{(i)}$ is net tile change at step $t$,
$\Delta\text{locks}_t^{(i)}$ counts only \emph{new} locks at step $t$,
and the terminal term delivers $+1.0$ for a win, $-1.0$ for a loss.

\section{Failure Mode Analysis}
\label{sec:failures}

We identify six compounding failure modes.
The first five are implementation bugs; the sixth emerges only after all five
are fixed, making it the subtlest and most instructive.

\subsection{Implementation Failure Modes (v1 Baseline)}
\label{sec:v1_failures}

Table~\ref{tab:failures} summarises each bug and its fix.

\begin{table}[t]
  \caption{Five implementation failure modes in the v1 PPO baseline and their
    targeted fixes applied in v2.}
  \label{tab:failures}
  \centering
  \renewcommand{\arraystretch}{1.3}
  \begin{tabularx}{\textwidth}{lXX}
    \toprule
    \textbf{Failure} & \textbf{v1 Bug} & \textbf{v2 Fix} \\
    \midrule
    No terminal reward
      & Win/loss gives zero reward; agent has no signal about the game outcome
      & $\pm 1.0$ terminal bonus on final step \\
    Reward-scale explosion
      & $+0.05 \times \text{total\_locks}$ per step (cumulative); episode
        returns reach $\pm$10{,}000
      & $+0.5 \times \text{new\_locks}$ per step; returns stay in $[+5, +30]$ \\
    Monte Carlo returns
      & MC over $T=250$ steps; step-1 advantages discounted by
        $0.99^{249} \approx 0.08$
      & GAE ($\lambda=0.95$); effective horizon $\approx 20$ steps \\
    Unnormalised observations
      & Raw positions $[0\text{--}9]$ mixed with steps remaining $[0\text{--}250]$
        in the same input layer
      & All inputs scaled to $[0,\,1]$ \\
    Wrong win detection
      & Winner determined by comparing cumulative episode rewards (corrupted
        by reward-scale bug)
      & Winner determined by counting tiles from the board observation \\
    \bottomrule
  \end{tabularx}
\end{table}

\paragraph{Reward-scale explosion.}
If an agent places 5 locks at step 50, the cumulative lock bug yields
$0.05 \times 5 \times 200 = 50$ additional reward from that decision alone,
completely dominating the tile-gain signal (+0.1 per tile) and the terminal
signal ($\pm 1$).
By episode 30{,}000, v1 episode returns ranged from $-10{,}000$ to
$+10{,}000$.
The v2 incremental fix restores meaningful scale: returns in $[+5, +30]$,
with the terminal reward providing a clear win/loss signal.

\paragraph{GAE vs.\ Monte Carlo returns.}
With $\gamma=0.99$ and $T=250$, the advantage of a step-1 action under
plain Monte Carlo returns is:
\[
  A_1^{\text{MC}} \propto \gamma^{T-1} = 0.99^{249} \approx 0.08
\]
The first $\sim50$ steps---where early territorial decisions are made---
contribute near-zero gradient signal.
GAE with $\lambda=0.95$ has effective credit-assignment horizon
$\nicefrac{1}{1-\lambda}=20$ steps, distributing gradient signal throughout
the episode.
Formally:
\begin{align}
  \delta_t   &= r_t + \gamma V(s_{t+1}) - V(s_t), \label{eq:td_error} \\
  \hat{A}_t  &= \sum_{k=0}^{T-t-1} (\gamma\lambda)^k \delta_{t+k}. \label{eq:gae}
\end{align}
Critic targets are $\hat{R}_t = \hat{A}_t + V(s_t)$ (TD-$\lambda$ returns).

\subsection{Competitive Overfitting: A Self-Play Failure Mode}
\label{sec:overfitting}

After fixing all five implementation bugs, PPO+GAE v2 achieves 73.5\% win
rate against a held-out random opponent at episode 8{,}000.
Continued training reveals a sharp reversal: by episode 12{,}000, win rate
collapses to 21.6\%---\emph{worse than the original broken v1 baseline}.

\paragraph{Mechanism.}
Because both agents train simultaneously, they co-adapt: Pink develops an
effective counter-strategy to Green's current policy, and Green simultaneously
counter-adapts.
Over thousands of episodes, each agent hyper-specialises against its
co-adaptive partner's \emph{specific} learned policy rather than against
general play.
The result is a Nash equilibrium between two narrowly-specialised policies
that generalises poorly to any out-of-distribution opponent---including the
uniformly-random agent.

\paragraph{Why standard monitoring fails.}
The self-play win rate (Pink's rolling-100 win rate against Green) remains
near 50\% throughout the collapse: both agents co-adapt equally, so the
dyad stays balanced.
The 51.9-percentage-point drop in generalisation win rate (73.5\% to 21.6\%) is
\emph{entirely invisible} unless the agent is periodically evaluated against
a fixed external baseline.
We therefore recommend periodic external evaluation as a standard diagnostic
in any self-play training regime.

\begin{figure}[t]
  \centering
  \realfig[7cm]{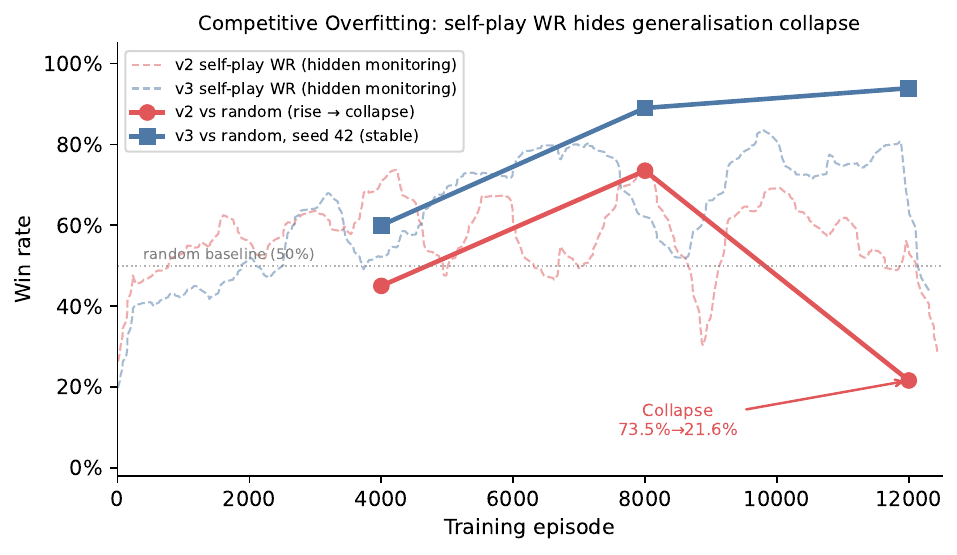}{%
    \textbf{Key figure.} Win rate vs.\ random agent over training episodes.
    v2: rises to 73.5\% at ep 8{,}000 then collapses to 21.6\% at ep 12{,}000.
    v3 (opponent mixing): stable throughout. Self-play WR stays near 50\% for
    both---showing the monitoring failure. Run: python3.9 make\_figures.py}
  \caption{Competitive overfitting in v2 self-play.
    Win rate vs.\ random rises then collapses, while self-play win rate gives
    no warning. Opponent mixing (v3) prevents the collapse.}
  \label{fig:overfitting}
\end{figure}

Figure~\ref{fig:overfitting} illustrates the v2 trajectory vs.\ the v3 fix.

\paragraph{Fix: Opponent Mixing.}
We mitigate competitive overfitting by replacing the co-adaptive Green agent
with a uniformly-random policy in fraction $\varepsilon$ of episodes:
\begin{equation}
  \pi_\text{Green}^{(t)} = \begin{cases}
    \text{Uniform}(\mathcal{A}) & \text{with probability } \varepsilon \\
    \pi_\text{Green}^\theta     & \text{with probability } 1-\varepsilon
  \end{cases}
\end{equation}
We use $\varepsilon=0.2$ (20\% random episodes), implemented as a single
conditional in the training loop.
This forces Pink to maintain strategies that generalise beyond the co-adaptive
policy, preventing over-specialisation.
Green continues to train on all episodes (including random-action episodes),
so no computational overhead is added.

\section{PPO+GAE Agent}
\label{sec:method}

\subsection{Network Architecture}
\label{sec:arch}

We use a shared-trunk actor-critic network:
\begin{itemize}
  \item \textbf{Trunk}: $206 \to 512$ (LayerNorm, ReLU)
    $\to 256$ (LayerNorm, ReLU) $\to 256$ (LayerNorm, ReLU)
  \item \textbf{Actor head}: $256 \to 128$ (ReLU) $\to 5$ (Softmax)
  \item \textbf{Critic head}: $256 \to 128$ (ReLU) $\to 1$
\end{itemize}
All linear layers use orthogonal initialisation (gain $\sqrt{2}$), except the
actor output (gain $0.01$, near-uniform initial policy) and critic output
(gain $1.0$).
LayerNorm replaces BatchNorm to handle the single-sample-per-update regime
of on-policy PPO.

\subsection{Observation Normalisation}
\label{sec:norm}

\begin{align*}
  \text{positions}_{[0:4]}       &\leftarrow \text{raw} / 9 \\
  \text{board state}_{[4:104]}   &\leftarrow \text{raw} / 2 \\
  \text{steps remaining}_{[204]} &\leftarrow \text{raw} / 250
\end{align*}
This maps all inputs to $[0,1]$, preventing position and step-count features
from numerically dominating board-state features during early gradient updates.

\subsection{Training Setup}
\label{sec:training}

Both agents train simultaneously with independent networks and optimisers
(fully decentralised).
Hyperparameters are given in Table~\ref{tab:hparams}.

\begin{table}[t]
  \caption{PPO+GAE v3 hyperparameters. Values are identical for both agents.}
  \label{tab:hparams}
  \centering
  \begin{tabular}{ll}
    \toprule
    \textbf{Hyperparameter}            & \textbf{Value} \\
    \midrule
    Discount $\gamma$                  & 0.99 \\
    GAE $\lambda$                      & 0.95 \\
    PPO clip $\epsilon$                & 0.2 \\
    Value loss coefficient             & 0.5 \\
    Entropy coefficient                & 0.01 \\
    PPO update epochs per episode      & 4 \\
    Learning rate                      & $3{\times}10^{-4}$ (Adam, $\varepsilon=10^{-5}$) \\
    Gradient clip norm                 & 0.5 \\
    Opponent mixing rate $\varepsilon$ & 0.2 \\
    Episode length                     & 250 steps \\
    Training episodes (per seed)        & 12{,}000 \\
    \bottomrule
  \end{tabular}
\end{table}

\section{Experiments}
\label{sec:experiments}

\subsection{Experimental Protocol}
\label{sec:protocol}

\begin{itemize}
  \item Each seed trains for 12{,}000 episodes with checkpoints every
    1{,}000 episodes.
  \item Generalisation win rate is measured at episode 12{,}000 by loading the
    checkpoint and running evaluation games against a uniformly-random opponent
    (no gradient updates during evaluation).
    Seeds 42--49: 100 games ($\pm 10$ pp Wilson CI);
    seeds 50--51: 500 games ($\pm 3.6$ pp Wilson CI).
  \item All runs tracked in Weights \& Biases, project
    \texttt{territory-paint-wars},\\
    group \texttt{ppo\_v3\_opponent\_mix}.
\end{itemize}

\subsection{10-Seed Convergence Study}
\label{sec:main}

Table~\ref{tab:main} reports results for all 10 seeds (42--51).

\begin{table}[t]
  \caption{%
    10-seed convergence study.
    \emph{Cumul.\ WR}: Pink cumulative self-play win rate at ep 12{,}000.
    \emph{vs.\ Random}: generalisation win rate against a uniformly-random
    opponent (100 games for seeds 42--49; 500 games for seeds 50--51).
  }
  \label{tab:main}
  \centering
  \begin{tabular}{ccc}
    \toprule
    \textbf{Seed} & \textbf{Cumul.\ WR (self-play)} & \textbf{vs.\ Random WR} \\
    \midrule
    42 & 64.0\%        & 93.9\% \\
    43 & 55.4\%        & 56.8\% \\
    44 & 49.9\%        & 84.8\% \\
    45 & 67.1\%        & 90.8\% \\
    46 & 56.1\%        & 55.7\% \\
    47 & 55.5\%        & 75.8\% \\
    48 & 55.6\%        & 80.6\% \\
    49 & 56.3\%        & 74.7\% \\
    50 & 59.0\%        & 79.2\% \\
    51 & 58.4\%        & 78.5\% \\
    \midrule
    \textbf{Mean $\pm$ std (10 seeds)}
       & ---
       & $77.1 \pm 12.6\%$ \\
    \bottomrule
  \end{tabular}
\end{table}

The range 55.7\%--93.9\% reflects genuine variation in convergence speed.
Seeds 43 and 46 are the weakest at ep 12{,}000; both were still improving at
the measurement point (their cumulative self-play WR had not plateaued),
suggesting slow convergence rather than a fundamentally worse policy.
The ep-12k evaluation point was chosen as the standardised cross-seed
comparison to prevent cherry-picking.

\paragraph{Statistical significance.}
Against the random-play null hypothesis ($\mu_0 = 50\%$), a one-sample
$t$-test gives $t(9) = 6.80$, $p < 0.001$, confirming that v3 agents
generalise well above chance.
Against the v2 ep-12k baseline ($\mu_0 = 21.6\%$), $t(9) = 13.93$,
$p < 10^{-7}$---the improvement from opponent mixing is not attributable
to variance across seeds.

\paragraph{Self-play WR as a generalisation proxy.}
Across the 10 seeds, the Pearson correlation between cumulative self-play WR
and vs-random WR is $r = 0.47$ (moderate positive), confirming that self-play
WR is a noisy but non-trivial predictor of generalisation \emph{within} v3.
This is in sharp contrast to v2, where self-play WR stayed near 50\% while
generalisation WR collapsed to 21.6\%---further evidence that self-play WR
is an inadequate monitoring signal without periodic external evaluation.

\begin{figure}[t]
  \centering
  \realfig[7cm]{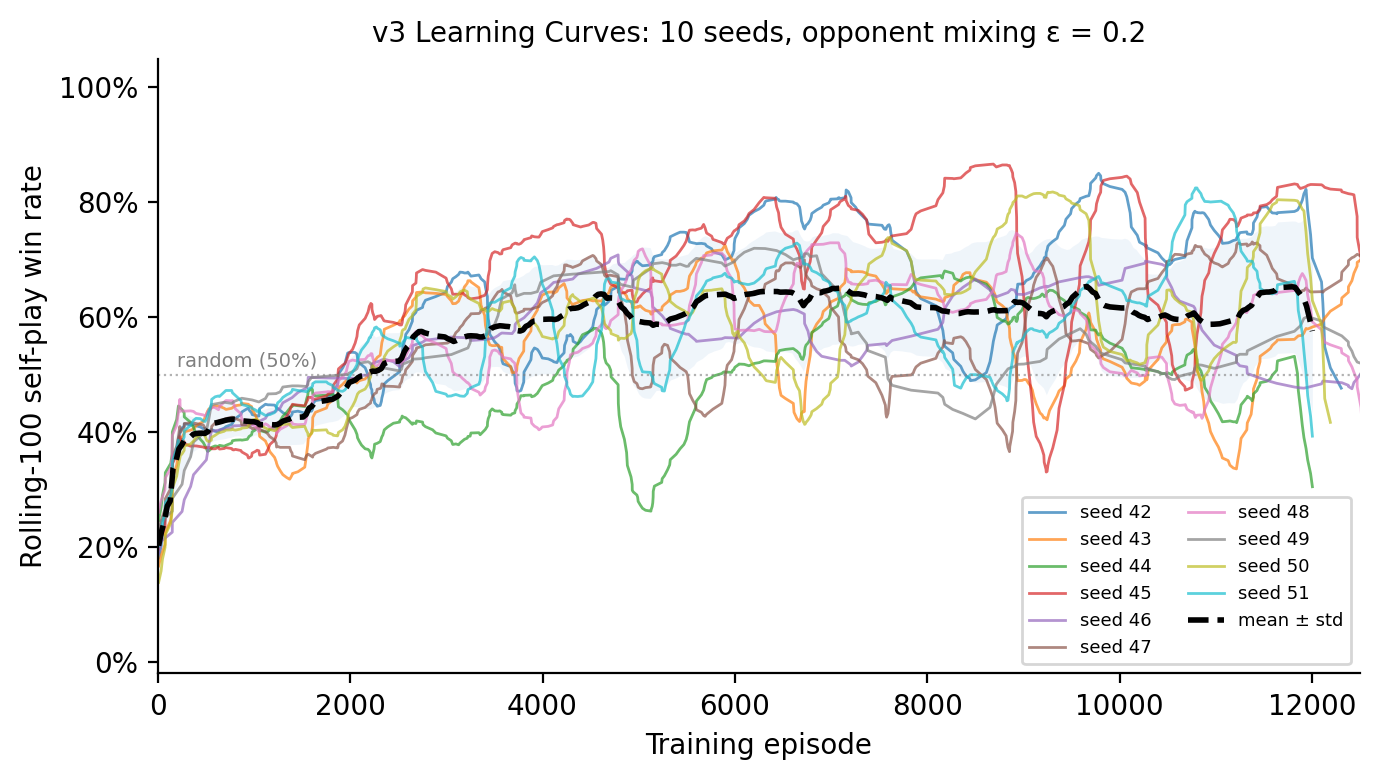}{%
    Rolling-100 self-play win rate for seeds 42--51 over 12{,}000 episodes.
    All seeds rise from $\sim$50\% with oscillation due to co-adaptation.
    Run: python3.9 make\_figures.py}
  \caption{Learning curves: rolling-100 Pink self-play win rate for all 10
    seeds (v3, opponent mixing). Despite high rolling variance (co-adaptive
    oscillation), the cumulative win rate climbs steadily for all seeds,
    confirming persistent learning.}
  \label{fig:learning_curves}
\end{figure}

Figure~\ref{fig:learning_curves} shows rolling-100 self-play win rates for
all completed seeds.

\subsection{Ablation Study}
\label{sec:ablation}

To isolate each fix's individual contribution, we train ablated variants
(seed 42, single run each) removing one fix at a time.
The ``$-$opponent mixing'' row uses the already-measured v2 result (21.6\%).

\begin{table}[t]
  \caption{%
    Ablation study. Each variant removes exactly one fix from v3.
    Win rate vs.\ random at episode 12{,}000 (seed 42 unless noted).
    $\Delta$ is computed relative to the 10-seed mean (77.1\%).
    $^\dagger$Seed 42 v3 full achieved 93.9\%; seed-matched $\Delta = -6.8$ pp.
  }
  \label{tab:ablation}
  \centering
  \begin{tabular}{lcc}
    \toprule
    \textbf{Variant}                   & \textbf{vs.\ Random WR} & \textbf{$\Delta$ from v3} \\
    \midrule
    v3 full (all fixes, 10-seed mean)  & 77.1\%        & --- \\
    \midrule
    $-$ terminal reward                & $87.1\%$      & $+10.0$ pp$^\dagger$ \\
    $-$ GAE (plain MC returns)         & $9.6\%$       & $-67.5$ pp \\
    $-$ obs.\ normalisation            & $12.6\%$      & $-64.5$ pp \\
    $-$ opponent mixing (= v2, ep 12k) & 21.6\%        & $-55.5$ pp \\
    \midrule
    v1 baseline (all bugs)             & 26.8\%        & $-50.3$ pp \\
    \bottomrule
  \end{tabular}
\end{table}

The ``$-$opponent mixing'' entry is directly measured from the v2 run at
episode 12{,}000, giving a clean 55.5-percentage-point attribution to opponent
mixing alone.

A striking pattern emerges: three of the four ablations produce an agent
\emph{worse than the broken v1 baseline} (26.8\%), all far below random
(50\%); the terminal-reward ablation is the exception, discussed below.

The ``$-$GAE'' agent wins only $9.6\%$ of games ($-67.5$ pp from v3).
With $\gamma^{249}\approx0.08$ discounting early-step advantages to near zero,
the agent receives no meaningful gradient signal about early territorial
decisions and learns an actively harmful late-game policy.

The ``$-$obs normalisation'' agent wins $12.6\%$ ($-64.5$ pp).
The training trajectory reveals why: the agent appeared to be learning
during self-play (rolling WR reached 97--100\% by episode 8{,}400), but this
was pure competitive overfitting---the same failure mode as v2, compressed
into fewer episodes because the misscaled inputs accelerate co-adaptive
specialisation.
Against a random opponent the policy collapses to $12.6\%$.
Obs normalisation thus serves a dual function: it enables stable gradient
updates \emph{and} slows the co-adaptive overfitting process enough for
generalised strategies to develop.

The ``$-$terminal reward'' result is the paper's most surprising finding:
removing the $\pm1.0$ terminal signal yields \textbf{87.1\%} vs.\ random
($\pm 3.0$ pp Wilson CI, 500 games)---above the 10-seed v3 mean (77.1\%)
and only $6.8$ pp below seed 42's own v3 result (93.9\%).
The agent learned a strong, generalisable territory-control policy
driven entirely by dense step rewards: tile gain ($+0.1$) and lock bonus
($+0.5$).
The terminal reward matters at the margin but is not individually necessary
when GAE, observation normalisation, and opponent mixing are all in place.

\subsection{Behavioural Analysis}
\label{sec:behaviour}

\begin{figure}[t]
  \centering
  \realfig[5.5cm]{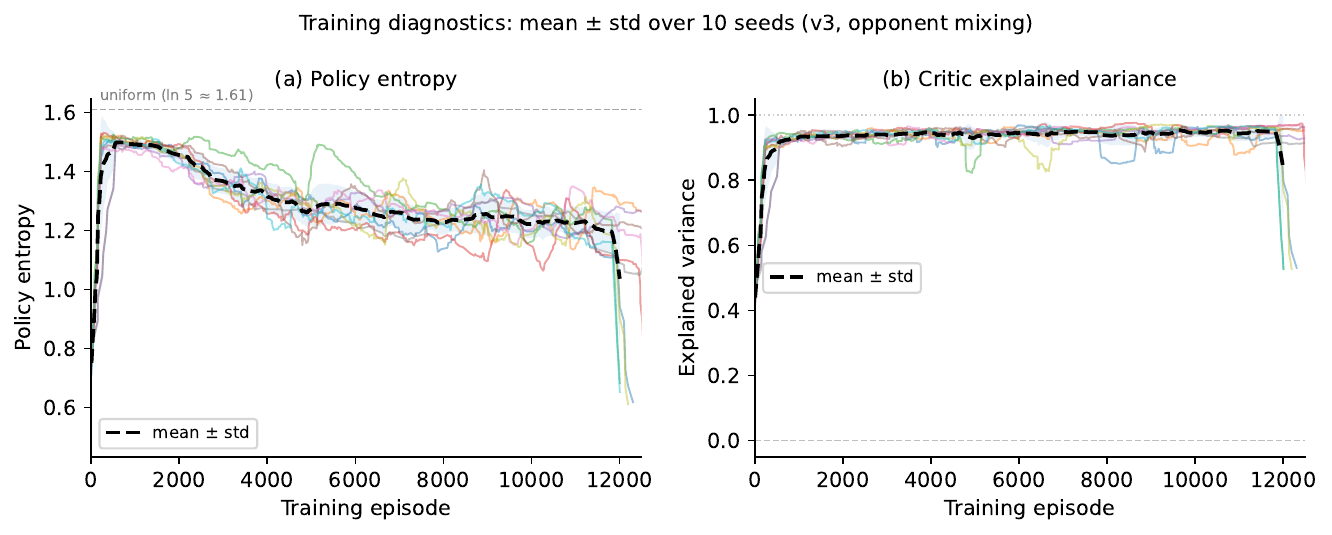}{%
    Policy entropy (left) and critic explained variance (right)
    over 12{,}000 episodes, mean $\pm$ std over 10 seeds.
    Run: python3.9 make\_figures.py}
  \caption{Training diagnostic curves (mean $\pm$ std over 10 seeds).
    Entropy decreases as the policy specialises; explained variance rises
    toward 1.0, confirming the critic learns accurate TD-$\lambda$ return
    estimates.}
  \label{fig:training}
\end{figure}

\paragraph{Training diagnostics.}
Figure~\ref{fig:training} shows policy entropy and critic explained variance
over training (mean $\pm$ std over 10 seeds).
Entropy starts at $\ln 5 \approx 1.61$ (uniform) and decreases as the
policy specialises.
Explained variance rises from $\approx 0$ toward values near 1.0, confirming
that the critic learns to accurately predict TD-$\lambda$ returns---a
necessary condition for GAE to provide useful advantage estimates.

\section{Discussion}
\label{sec:discussion}

\paragraph{Competitive overfitting as an underreported failure mode.}
The 51.9-percentage-point collapse in generalisation win rate is entirely invisible to
standard self-play monitoring: the self-play win rate fluctuates near 50\%
throughout because both agents co-adapt equally.
The failure is revealed only by periodic evaluation against a fixed external
baseline---a diagnostic step that is easy to add but rarely reported.
We recommend it as standard practice in any self-play training regime.

\paragraph{Opponent mixing as a minimal fix.}
The fix is deliberately minimal: one conditional in the training loop,
no additional networks, no curriculum, no memory overhead.
This distinguishes it from population-based training
\citep{bansal2018emergent} and league play \citep{vinyals2019grandmaster},
which require maintaining multiple agents simultaneously.
The simplicity is intentional---it isolates the mechanism (maintaining
exposure to diverse opponents) from implementation complexity.

Opponent mixing can be interpreted as preventing overfitting to the narrow
empirical game distribution induced by pure self-play.
By injecting a fixed stochastic policy into a fraction of episodes, the
training distribution is broadened, encouraging policies that perform well
across a wider support of opponent behaviours.
This is analogous to regularisation in supervised learning, where exposure
to diverse training inputs improves generalisation to unseen examples.
The random opponent acts as an implicit diversity constraint, preventing the
co-adaptive dyad from collapsing into a mutually specialised equilibrium.

\paragraph{Why GAE is critical, not merely a convenience.}
In single-agent tasks with short episodes, MC returns are often adequate.
In a 250-step competitive game, early territorial decisions (first $\sim$30
steps) largely determine the outcome, yet they receive gradient weight
$0.99^{249} \approx 0.08$ under MC.
The ablation makes this concrete: removing GAE ($\lambda=1.0$, plain MC
returns) produces an agent that wins only $9.6\%$ of games against a
uniformly-random opponent after 12{,}000 episodes of training---$-67.5$ pp
below v3 and below the broken v1 baseline.
The agent does not merely fail to improve; it learns an actively harmful
policy, presumably overfitting to late-game cues that carry negligible
gradient weight at the relevant decision points.
GAE with $\lambda=0.95$ is not merely an optimisation in this setting; it is
critical for the agent to receive any meaningful signal about its early-game
strategy.

\paragraph{Observation normalisation has a hidden second role.}
Removing obs normalisation does not merely slow learning---the agent reaches
97--100\% rolling self-play win rate by episode 8{,}400, superficially
suggesting strong performance.
But the eval result is $12.6\%$ vs.\ random: the agent co-adaptively
overfitted faster than v2, collapsing even harder.
The misscaled inputs ($[0\text{--}9]$ positions mixed with $[0\text{--}250]$
steps remaining) appear to bias the network toward large-magnitude features,
accelerating co-adaptive specialisation at the expense of generalisation.
Obs normalisation thus serves a dual purpose: gradient stability \emph{and}
implicit regularisation against competitive overfitting.

\paragraph{Terminal reward is complementary, not individually necessary.}
Removing the $\pm1.0$ terminal win/loss signal yields 87.1\% vs.\ random
(seed 42)---the strongest single-seed ablation result and above the
10-seed v3 mean.
This is a discriminative finding: it shows that the dense step-reward
signal (tile gain + lock bonus) is sufficient for learning a strong
generalisable policy when the other three fixes are in place.
Terminal reward matters at the margin ($-6.8$ pp seed-matched) but is not
a prerequisite.
By contrast, GAE, observation normalisation, and opponent mixing are each
individually critical---their removal collapses performance to below the
broken v1 baseline.
This asymmetry reveals that not all fixes contribute equally:
some address catastrophic failure modes that eliminate learning
(GAE and obs normalisation for credit assignment; opponent mixing for
co-adaptation), while terminal reward sharpens the win signal but is not
a prerequisite when dense step rewards are present.

\paragraph{Reward scale matters independently of reward sign.}
The cumulative lock bug had the correct sign (more locks = more reward) but
catastrophically wrong scale.
This is a less-studied failure mode than reward misspecification; it arises
whenever an environment accumulates statistics and reports them as per-step
bonuses.
Practitioners should audit not only reward \emph{sign} but reward
\emph{scale} relative to other signal components.

\paragraph{Limitations.}
Results are limited to a single environment and algorithm.
The relative importance of each fix may differ in shorter-episode or
sparser-reward settings.
The Unity--Python TCP bridge introduces latency ($\sim$35 episodes/minute
on an M-series Mac), limiting experiment scale.
Seeds 42--49 were evaluated at 100 games ($\pm 10$ pp Wilson CI); seeds 50--51
at 500 games ($\pm 3.6$ pp Wilson CI).
We do not evaluate whether these findings extend to larger-scale or stochastic
environments, which we leave for future work.

\section{Conclusion}
\label{sec:conclusion}

We present a systematic case study of PPO failure in competitive multi-agent
RL.
Starting from a baseline that performs worse than random (26.8\% win rate),
we identify and fix five implementation failure modes, then identify a
sixth---competitive overfitting---that is unique to self-play and invisible
to standard win-rate monitoring.
Fixing competitive overfitting with a one-line opponent mixing change
recovers generalisation win rate from 21.6\% to $77.1\%$ ($\pm 12.6\%$,
10 seeds), with individual seeds reaching 93.9\%.
Ablation experiments yield a discriminative result.
GAE, observation normalisation, and opponent mixing are each individually
critical: removing any one collapses win rate to below the broken v1
baseline (26.8\%)---to $9.6\%$, $12.6\%$, and $21.6\%$ respectively.
Terminal reward, by contrast, is complementary rather than necessary:
removing it yields $87.1\%$ vs.\ random, only $6.8$ pp below the seed-42
v3 result, confirming that the dense step-reward signal alone is
sufficient when the other three fixes are in place.
This asymmetry---three catastrophic ablations, one benign---shows that the
v1 failure modes target distinct mechanisms and do not all compound
equally.

We open-source \TPW{} as a competitive RL benchmark and hope the
failure-mode taxonomy aids practitioners debugging similar training failures.
The competitive overfitting finding---that self-play win rates can hide
generalisation collapse---is the primary empirical contribution and applies
broadly to any self-play regime.

\section*{Acknowledgments}
The author is a high school student at West Windsor-Plainsboro High School South
conducting this research independently.
All experiments were run on personal hardware (Apple M-series).
No external funding was received.

\bibliographystyle{abbrvnat}
\bibliography{references}

\appendix
\section{Environment Implementation Details}
\label{app:unity}

The Unity scene contains two \texttt{AgentController} components,
\texttt{GridManager}, \texttt{GameManager}, and \texttt{PythonBridge}.
\texttt{PythonBridge} runs a background thread that accepts one TCP connection
and exchanges JSON messages synchronously with the Python training loop.

Message protocol:
\begin{itemize}
  \item \textbf{Python $\to$ Unity (action)}: \texttt{\{"pink\_action": int, "green\_action": int\}}
  \item \textbf{Unity $\to$ Python (step)}:
    \texttt{\{"pink\_obs": [\ldots], "green\_obs": [\ldots],}\\
    \texttt{"pink\_reward": float, "green\_reward": float, "done": bool\}}
  \item \textbf{Reset}: Python sends action $= -1$; Unity resets and returns initial state.
  \item \textbf{Position swap}: Python sends action $= -2$; Unity swaps starting positions (used for position-bias validation).
\end{itemize}

\section{Ablation Commands}
\label{app:ablation_commands}

Dedicated ablation scripts are provided in \texttt{Python/}.
Each script is a self-contained copy of \texttt{train\_v2\_gae.py} with
exactly one fix removed; all other hyperparameters are identical.

\begin{itemize}
  \item \textbf{Remove opponent mixing} (already measured --- v2 ep 12k = 21.6\%):
    \texttt{python3.9 train\_v2\_gae.py --seed 42 --opponent-random 0.0}
  \item \textbf{Remove GAE} ($\lambda=1.0$, plain Monte Carlo returns):
    \texttt{python3.9 ablation\_no\_gae.py --seed 42 --episodes 12000}\\
    (Sets \texttt{self.lam=1.0} in \texttt{PPOAgent}; \texttt{lam=1.0}
    recovers MC returns, \texttt{lam=0.0} would give 1-step TD---a different
    ablation.)
  \item \textbf{Remove terminal reward} (subtract Unity's $\pm1.0$ terminal
    component on the Python side when \texttt{done=True}):
    \texttt{python3.9 ablation\_no\_terminal\_reward.py}\\
    \texttt{--seed 42 --episodes 12000}
  \item \textbf{Remove obs normalisation} (raw observations passed directly
    to network):
    \texttt{python3.9 ablation\_no\_obs\_norm.py --seed 42 --episodes 12000}
\end{itemize}

After each training run, evaluate the saved \texttt{\_final.pth} checkpoint
against 500 random-opponent games (replace \texttt{<name>} with the ablation
script prefix):
\texttt{python3.9 eval\_500.py <name>\_seed42\_final.pth --games 500}

\section{Reproducibility Checklist}
\label{app:repro}

\begin{itemize}
  \item Seeds reported: 42--51 (all complete).
  \item Unity version: 2022.3 LTS.
  \item Python: 3.9. PyTorch: 2.2.2.
  \item Training: \texttt{python3.9 train\_v2\_gae.py --seed \{N\}}
  \item Checkpoints: \texttt{pink\_v3\_checkpoint\_seed\{N\}\_\{ep\}.pth}
  \item Evaluation (500 games, Wilson CI): \texttt{python3.9 eval\_500.py}\\
    \texttt{\hspace*{1em}pink\_v3\_checkpoint\_seed\{N\}\_12000.pth --games 500}
  \item WandB runs:
    \href{https://wandb.ai/singhdiyansha-west-windsor-plainsboro-high-school-south/territory-paint-wars}{wandb.ai $\to$ territory-paint-wars},
    group \texttt{ppo\_v3\_opponent\_mix}.
\end{itemize}

\end{document}